# Principles for Evaluation of AI/ML Model Performance and Robustness

O.M. Brown
A.B. Curtis
J.A. Goodwin

21 January 2021
Revised 30 March 2021

## Lincoln Laboratory

MASSACHUSETTS INSTITUTE OF TECHNOLOGY

*LEXINGTON, MASSACHUSETTS*

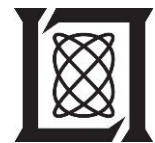

This material is based upon work supported by the Under Secretary of Defense for Research and Engineering under Air Force Contract No. FA8702-15-D-0001.

This report is the result of studies performed at Lincoln Laboratory, a federally funded research and development center operated by Massachusetts Institute of Technology. This material is based upon work supported by the Under Secretary of Defense for Research and Engineering under Air Force Contract No. FA8702-15-D-0001. Any opinions, findings, conclusions or recommendations expressed in this material are those of the author(s) and do not necessarily reflect the views of the Under Secretary of Defense for Research and Engineering.

# Massachusetts Institute of Technology
# Lincoln Laboratory

## Principles for Evaluation of AI/ML Model Performance and Robustness

*O.M. Brown*
*Group 01*

*A.B. Curtis*
*J.A. Goodwin*
*Group 36*

**Lexington**                                                    **Massachusetts**

This page intentionally left blank.

# EXECUTIVE SUMMARY

The Department of Defense (DoD) has significantly increased its investment in the design, evaluation, and deployment of Artificial Intelligence and Machine Learning (AI/ML) capabilities to address national security needs [1, 2]. While there are numerous AI/ML successes in the academic and commercial sectors, many of these systems have also been shown to be brittle and nonrobust [3]. In a complex and ever-changing national security environment, it is vital that the DoD establish a sound and methodical process to evaluate the performance and robustness of AI/ML models before these new capabilities are deployed to the field [4].

Without an effective evaluation process, the DoD may deploy AI/ML models that are assumed to be effective given limited evaluation metrics but actually have poor performance and robustness on operational data. Poor evaluation practices lead to loss of trust in AI/ML systems by model operators and more frequent—often costly—design updates needed to address the evolving security environment. In contrast, an effective evaluation process can drive the design of more resilient capabilities, flag potential limitations of models before they are deployed, and build operator trust in AI/ML systems.

This paper reviews the AI/ML development process, highlights common best practices for AI/ML model evaluation, and makes the following recommendations to DoD evaluators to ensure the deployment of robust AI/ML capabilities for national security needs:

- Develop testing datasets with sufficient variation and number of samples to effectively measure the expected performance of the AI/ML model on future (unseen) data once deployed,

- Maintain separation between data used for design and evaluation (i.e., the test data is not used to design the AI/ML model or train its parameters) in order to ensure an honest and unbiased assessment of the model's capability,

- Evaluate performance given small perturbations and corruptions to data inputs to assess the smoothness of the AI/ML model and identify potential vulnerabilities, and

- Evaluate performance on samples from data distributions that are shifted from the assumed distribution that was used to design the AI/ML model to assess how the model may perform on operational data that may differ from the training data.

By following the recommendations for evaluation presented in this paper, the DoD can fully take advantage of the AI/ML revolution, delivering robust capabilities that maintain operational feasibility over longer periods of time, and increase warfighter confidence in AI/ML systems.

This page intentionally left blank.

# TABLE OF CONTENTS

This page intentionally left blank.

# LIST OF FIGURES

This page intentionally left blank.

# 1. INTRODUCTION

Recent years have seen major academic and industry breakthroughs in the field of Artificial Intelligence and Machine Learning (AI/ML), and the Department of Defense (DoD) has identified the need to harness these advances to improve our national security and maintain a strategic advantage [2]. Despite numerous successes in AI/ML, many of these new technologies have been shown to be brittle, uninterpretable, and easily fooled [3]. The DoD recognizes the need to improve existing evaluation practices to measure and improve the resiliency, robustness, and reliability of AI/ML models before such models are deployed as an operational capability [1,4].

We share the urgency to improve DoD AI/ML evaluation processes. We have observed that in many cases, the defense community has not adhered to best practices from the AI/ML community in evaluating model performance and robustness. Without proper evaluation practices, understanding of the capabilities of AI/ML models is biased, often assuming a larger capability than is warranted. Such a mismatch between predicted and operational performance can lead to warfighter distrust of AI/ML systems. Additionally, poor evaluation practices may in turn be incentivizing poor design approaches. For example, if the primary metric for evaluation is accuracy on a known set of testing data, a solution that memorizes that data may be selected over one that works against the full range of potential real-world variations of that data.

In this paper, we discuss the AI/ML development process and provide recommendations on how the DoD can encourage the design of robust AI/ML models through its evaluation procedures. To be considered robust, an AI/ML model should exhibit the following properties:

1. **Generalization:** An AI/ML model should be able to achieve good performance on data that is drawn from the same distribution as the training data but not used explicitly during training (i.e., generalize to unseen data).

2. **Robustness:** An AI/ML model should maintain performance, with graceful degradation, as the unseen test data becomes increasingly different from the training data due to:

   (a) **Sample Uncertainty:** small perturbations or corruptions applied to the data, and

   (b) **Distribution Uncertainty:** variations between the expected and observed distributions of the data, e.g., due to unanticipated changes in the scenario once deployed or the evolution of threat capabilities.

An AI/ML model that exhibits these properties will perform well against expectations while being robust to unexpected variations once deployed operationally. Other concepts such as the ability to detect outliers and the ability to autonomously adapt once the model is deployed are also important for AI/ML model robustness, but are out of scope for this paper. Note that these properties apply whether we are assessing robustness of an entire AI/ML system involving many components, or a single component of that system that applies AI/ML. We adopt common terminology in referring to this system or algorithm as an AI/ML model, and provide some tools in this paper for model evaluators to apply in assessing the quality of these models.

## 2. STANDARD AI/ML EVALUATION PRACTICES

The goal in a supervised AI/ML problem is to take a set of data examples containing labeled input-output pairs (e.g., images to class labels, audio signals to text transcriptions) that have been sampled from some data distribution and design an AI/ML model (e.g., neural network, decision tree) that can predict outputs given the inputs. The parameters of the model (e.g., weights of neural network, thresholds of decision tree) are typically learned through an optimization procedure using this set of *training data*. Note that the data inputs are often measurements collected by a sensor (e.g., camera, microphone), and the observations may contain both the signal of interest as well as other background signals and noise. An AI/ML model must learn to extract the relevant information from the noisy observations.

### 2.1 TRAINING-VALIDATION-TEST PARADIGM

Ideally, the learned AI/ML model is a close approximation of the true model from which the data is generated. The true model is almost always unknown, but given sample points (input-output pairs) in the training set, the predicted output by the learned model is compared with the true output for each sample to compute an estimate of the model's performance. The error rate on the training set is called the *in-sample error* or *training error*. It is this error rate (or a similar measure computed on the training set) that is directly minimized via the optimization procedure.

Matching the true model on this training set is not enough; the real objective is for the learned model to match the true model across the entire data distribution, so that it performs well on any operational data encountered once the model is deployed. The goal is to minimize the error rate on the full data distribution, called the *out-of-sample error*; however, this error rate cannot be computed via a fixed set of samples. Furthermore, since the model and its parameters can be adjusted based on observed performance on the training data, the measured in-sample error is often biased and does not reflect the expected performance on unseen data. Efforts to reduce the error on the training set may even increase the out-of-sample error, a phenomenon referred to as overfitting.

The out-of-sample error can instead be estimated (before the model is deployed) with the use of *test data*, which is a set of samples that are drawn from the data distribution but are not part of the training set or used in any portion of the training process. The *test error* is the error rate on these previously-unseen data samples. The training and testing sets are both subject to sampling bias, since the true data distribution may not be completely known. As long as the test set is not also included in the model training process, however, there is no additional bias when estimating the out-of-sample error with a test set.

Estimating the out-of-sample error is important, but the ultimate goal is to minimize this value. Knowing the perils of using the training set performance to solely predict an AI/ML model's ability to generalize to unseen data, it is common to withhold a portion of the training data (typically 10-20%), called the *validation data*, from the training process. The learning algorithm uses the training data to search over the hypothesis space of models that minimize the *validation error*. Instead of selecting a fixed validation set, the model developer may choose to use an approach called *cross-validation*, where the training set is repeatedly split into different disjoint training-validation subsets, and average validation error across these subsets is used as a prediction for the test error. These approaches are useful for training models that are able to generalize to new data rather than memorize the training set.

This *training-validation-test paradigm* is a practice that is standard in the AI/ML community and should be fully embraced by the DoD. Additional practices are beginning to emerge in the AI/ML community that are aimed at evaluating model robustness to small perturbations of the data samples and larger shifts of the data distribution. These practices will also be useful for DoD to evaluate AI/ML model robustness, and are discussed further in Sections 3.3 and 3.4.

## 2.2  THE AI/ML DEVELOPMENT PROCESS

In Figure 1, we illustrate the training-validation-test paradigm within the AI/ML development process, which consists of three phases: design, evaluation, and deployment. During the *design phase*, the model developer designs an AI/ML model aimed at meeting the requirements established and verified by a separate model evaluator during the *evaluation phase*. The model developer uses training and validation data to design the model and predict its performance while the model evaluator verifies performance using a set of test data that is unseen by the developer. After a model is evaluated, the developer may have the opportunity to make additional updates to the model, after which it undergoes a new evaluation. This process is repeated until the evaluator is satisfied with the performance, at which point the model enters the *deployment phase*, where it becomes operational. The performance of the deployed model on operational data is then monitored by model operators, and further design updates and subsequent evaluations are conducted when deemed necessary.

## 2.3  AI/ML DEVELOPMENT AND EVALUATION IN THE DoD

In the defense industry, the model operator is the end user of the AI/ML system (e.g., military service members in the field). The model developer is often a defense contractor, a government research lab, or a team within the government itself. In some programs, there may even be multiple groups competing to develop the best model according to the evaluation criteria. The model evaluator is often a government team or an outside independent entity. The model evaluator's goal is to determine the effectivity of the model against the types of data the model may encounter when it is deployed. The model developer has the goal of designing a model that performs well when deployed, but is very strongly motivated by the need to meet the evaluation criteria set forth by the evaluator. Though the developer may provide its own assessment of model performance, this should not replace the verification done by the model evaluator. A knowledgeable and independent model evaluator plays a vital role in ensuring the robustness of the AI/ML system.

The remainder of this paper focuses on the evaluation stage of the development process. DoD model evaluators must recognize the ways in which assumptions made during the evaluation phase can influence the design phase for better or for worse. Such assumptions include the expected properties of the unseen data or specific requirements and scoring metrics that favor certain types of solutions. While model developers have the ultimate responsibility of designing an AI/ML model to solve a particular problem for the model operators, the model evaluator can establish an evaluation process that motivates good design practices by the developers and increases the chance of a robust model being deployed. Additionally, model evaluators must clearly communicate any assumptions made during the evaluation process to model operators so that warfighters know under which conditions a model has been tested and is thus expected to have performance.

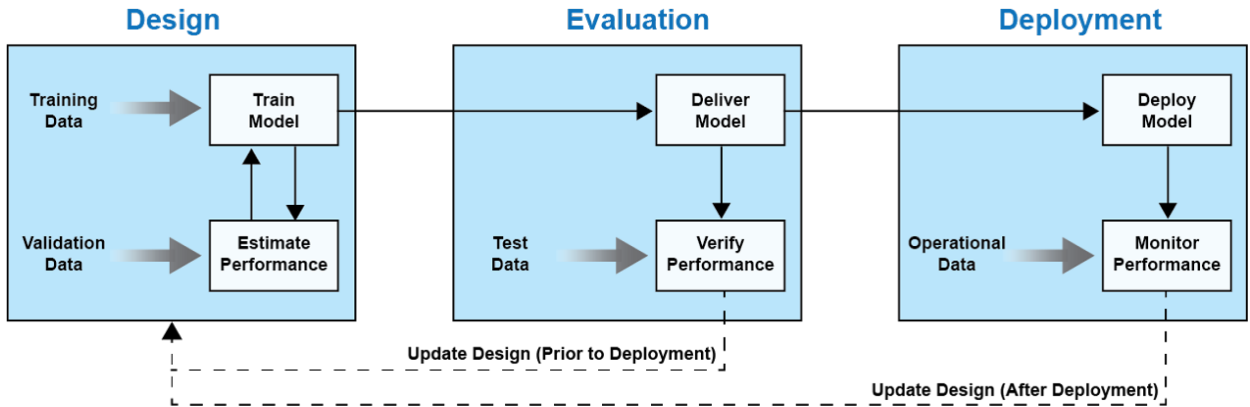

*Figure 1. AI/ML model development process. (Left) Design Phase: model developers use training data to optimize model parameters and validation data to estimate performance. (Middle) Evaluation Phase: model evaluators verify performance and robustness of the delivered model using test data that was not seen by developers during the training process. (Right) Deployment Phase: model transitions to an operational capability and its performance on operational data is monitored by model operators. Updates to the design are made when the model is not performing as desired in either the evaluation or deployment phase.*

It is easy, especially when faced with concerns about cost and schedule, to take shortcuts in the creation of test sets and the establishment and verification of requirements. It is critical for the DoD to realize that poor AI/ML evaluation processes may have significant negative effects on the robustness of a deployed model. We encourage model evaluators in the DoD to carefully consider their current assumptions and practices, and make improvements where necessary to ensure the robustness properties described in the intro of this paper are being properly evaluated and improved.

# 3. PRINCIPLES AND PRACTICES FOR DoD AI/ML MODEL EVALUATORS

In the following section, we present a set of principles and practices for model evaluators to appropriately assess and motivate the design of robust AI/ML models. These practices include selecting a representative data distribution, evaluating performance on unseen data, evaluating robustness to perturbations, and evaluating robustness to shifts in the data distribution. Model evaluators in the defense industry are encouraged to follow the lead of the AI/ML community in applying these evaluation practices.

## 3.1 SELECTING A REPRESENTATIVE DATA DISTRIBUTION

The purpose of the test set is to assess the performance of the model against unseen data and to provide an estimate of the out-of-sample error. Given that the true data distribution is often unknown and information about the make-up of the distribution is limited due to the nature of the contested security environments for the DoD, the following properties of a test set are necessary to provide a trustworthy estimate of the out-of-sample error:

- **Sample Diversity and Distribution**: In order to achieve an unbiased estimate of the out-of-sample error, the model evaluator should develop a test set that represents a diverse set of the important characteristics of the data distribution and the likelihood of such characteristics being realized. This may include potential physical characteristics of targets of interest (e.g., size, color, material), characteristics of the sensor (e.g., location, bandwidth, sensitivity), and environmental conditions (e.g., background clutter, time of day, weather). The actual range and distribution of potential inputs may not be fully known, and the bias from poorly estimating either of these components is difficult to quantify. It is important to reassess and adjust test set distributions if understanding and/or expectations about the real world threat change. Additionally, performance can be measured on different subsets of the full test set in order to gain insight into performance under different input conditions, even though the relative likelihood of those conditions may be uncertain. These efforts will allow the test error to be a more reliable and less biased indication of how the model will perform once deployed.

- **Sample Size**: Even if the test set distribution were perfectly known, model performance on two different test sets drawn from that distribution would not be expected to be identical. The test dataset must be large enough that the expected fluctuations in performance due to random sampling are sufficiently small, providing confidence that the estimated error represents the expected performance once the model is deployed. Large fluctuations of the performance error on new sets of test data might be an indication that the test dataset is too small or that the model is not robust.

Figure 2 shows an example of how sample diversity and size relate with regard to estimating the true out-of-sample error. Poor sampling (shown in orange and purple on the top and bottom, respectively) leads to large biases in the estimate of the true error, while improved sampling across the data distribution (shown in blue in the middle) lowers that bias. Increasing the number of samples in the test set reduces the variance (shaded regions) around the mean of the estimate (solid lines).

Improving data variability, distribution, and sample size might take the form of gathering and labeling more real data across a wider variety of environmental conditions, consulting with experts and reviewing intelligence to determine the likelihood of observing various inputs, or generating more simulation data that represent additional potential scenarios. The cost associated with collecting or generating more data samples and evaluating performance on these larger datasets is application-dependent, and in some situations this process has been considered cost-prohibitive. Given improvements in compute power, the ability to develop large datasets and analyze performance on those datasets has increased over time and will continue to increase. A balance always needs to be found between the added cost and the potential benefit of better performance, lower uncertainty, and more robustness.

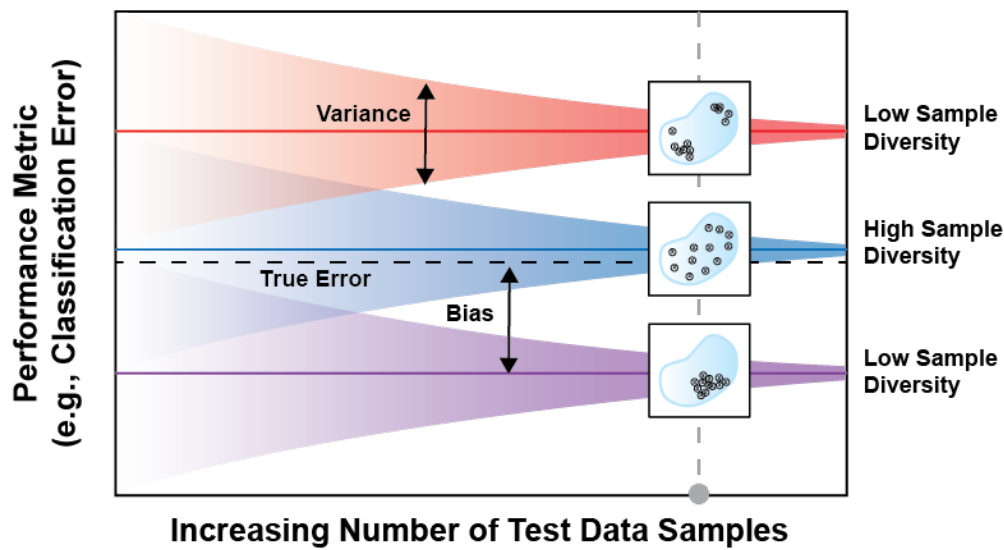

*Figure 2. Illustration of how sample diversity and sample size impact the bias and variance of the estimated out-of-sample performance metric. If the data distribution for the test set is sampled with lower diversity (top/orange and bottom/purple), the performance metric is likely to be biased away from the true error in an unknown direction. More sample diversity (middle/blue) reduces this bias. Increasing the number of test data samples reduces the variance (shaded regions) on the estimate of the performance metric (solid lines).*

## 3.2 EVALUATION OF GENERALIZATION CAPABILITY

In the training-validation-test paradigm that is ubiquitous in AI/ML model design, separating training data from testing data is fundamental. In some DoD applications, the cost of collecting or generating data can be high, and it can be tempting to ignore this principle and use all available data for training. However, if test data is provided and used during training, the reported performance is a biased estimate of the true out-of-sample error rate. In this situation, the generalization error appears artificially low, leading to a misunderstanding about the actual capability of the model and loss of operator trust.

Requiring performance against a test set that is truly unseen by the model developer encourages a design approach that is not solely focused on optimization against the training set. It motivates the use of techniques that both improve test set performance and help to predict that performance, such as validation, cross-validation, and variants of cross-validation where entire categories of the training data are held out instead of random subsets. These improved, stable AI/ML models reduce risk for model developers in the evaluation process and help model evaluators and operators build confidence in system performance metrics. Evaluators should also encourage model developers to report on their predictions of model performance during the design phase, even if there are no specific requirements for their predicted test set error to be close to the actual test set error.

Note that the specific mechanics of performing the evaluation may impact future evaluations. If a model developer is ever provided the test data in order to perform their own evaluation and report performance, that data is no longer "unseen" and should not be used for future tests. If a test set is used to optimize model parameters in any way, it is considered training data and reintroduces bias into the estimate of out-of-sample error. Whenever possible, such a situation should be avoided. The model evaluator may trust the model developer to not use test data for anything but the evaluation, but a better practice is to provide a completely new test set for evaluation at the next test event or program milestone. In some cases, it may be possible for the evaluator to provide the model developer with an unlabeled test set, and have the developer return a set of predicted labels to the evaluator for scoring. Alternatively, the model evaluator may require a deliverable model, allowing them to perform the evaluation without sharing the test data with the developer, making future reuse of that test set possible.

Figure 3 illustrates how generalization properties can be measured during the design and evaluation stages. Here, the model developer evaluates training (green) and validation (purple) errors across different model architectures of varying complexity (e.g., number of parameters or neural network layers) in order to select the model with the smallest validation error. The model evaluator then uses the delivered model to measure performance at a specific point along the test error curve (yellow line), which is an estimate of the true unknown error curve (black dashed line). A model based on minimizing the validation error (blue circle) is likely to be more robust and perform better against unseen data than a model based on minimizing the training error (red circle).

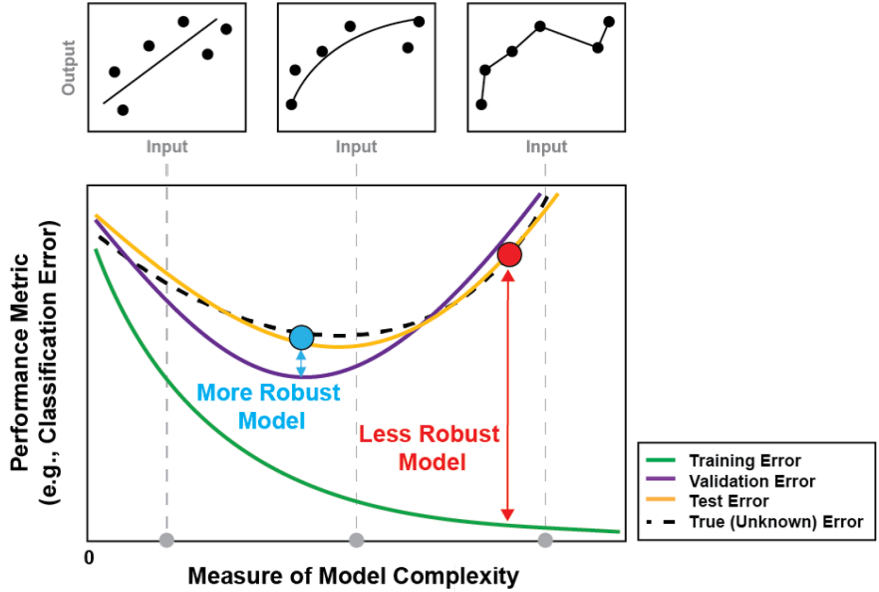

*Figure 3. Notional performance as a function of model complexity (e.g., number of neural network layers, number of features used in a decision tree). The model developer uses the training and validation data to select a model with the appropriate complexity to perform well against unseen data. The model evaluator then tests this model to determine the test error (yellow line), which is an estimate of the true unknown error (black dashed line). To minimize this true error, the model complexity should not be too low (underfitting) or too high (overfitting). Since training error (green line) almost always decreases as model complexity increases, a model chosen based on minimizing this error will tend to be less robust when testing on unseen data (red circle). The validation error curve (purple line) more closely tracks the true error, and a model chosen based on minimizing this error will likely be more robust (blue circle).*

### 3.3 EVALUATION OF ROBUSTNESS TO SAMPLE UNCERTAINTY

The AI/ML community has found that evaluating performance on a withheld test set is not sufficient for fully characterizing a model's performance against unseen data. Many AI/ML models have been shown to suffer significant drops in performance when presented with slightly perturbed data inputs due to common corruptions such as blurring in images [5], background noise in audio [6], and targeted manipulation of the inputs by an optimization procedure, often referred to as adversarial examples [3].

A robust AI/ML model should be able to maintain performance in the presence of small changes in the data samples and performance should gracefully degrade as those changes become larger. In other words, if two inputs are considered close for a given domain, their corresponding outputs produced by the model should also be close. This property is sometimes referred to as smoothness, stability, or robustness to perturbations; in this paper, we refer to it as robustness to sample uncertainty.

Refer to Figure 4 for a depiction of this property. In this example, the less robust model incorrectly classifies the image after the background pixels of the image have been perturbed by a small amount. In contrast, the more robust model has learned a different feature representation of the input space such that this small perturbation does not cause the sample to be classified incorrectly. AI/ML models that are robust to sample uncertainty are not easily fooled by adversarial examples and do not suffer significant performance loss in the face of common input corruptions.

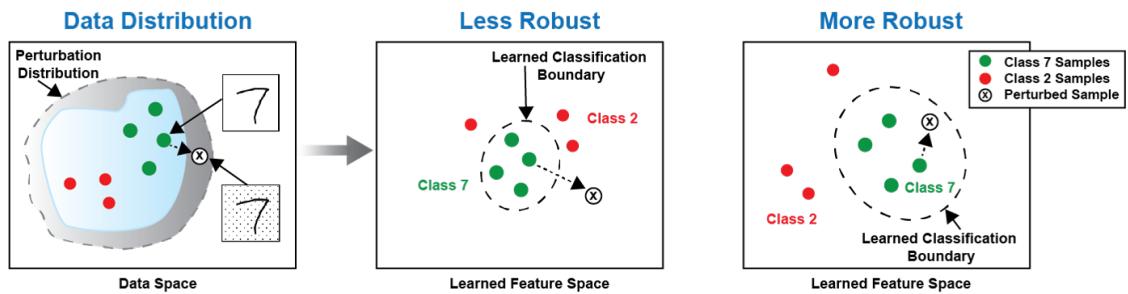

*Figure 4. Notional depiction of models that are more robust (right) and less robust (middle) to sample uncertainty, which is represented as a small perturbation to the pixels of the image of the number 7. The less robust model has learned a feature space and classification boundary where a small perturbation in the data space corresponds to a large perturbation in the feature space, causing the perturbed sample to be misclassified. The more robust model instead learns a feature representation and classification boundary that tolerates this perturbation.*

To assess robustness to sample uncertainty, we recommend creating robustness curves, which plot a model's performance as a function of perturbation size. See Figure 5 for an example. A robust model has graceful degradation in performance as the size of perturbations applied to the testing samples is increased (blue curve in the figure). While such a model may sacrifice lower error on the original test set (distance = 0), it is preferred to a model where error rates quickly increase when encountering data that is slightly perturbed (red curve).

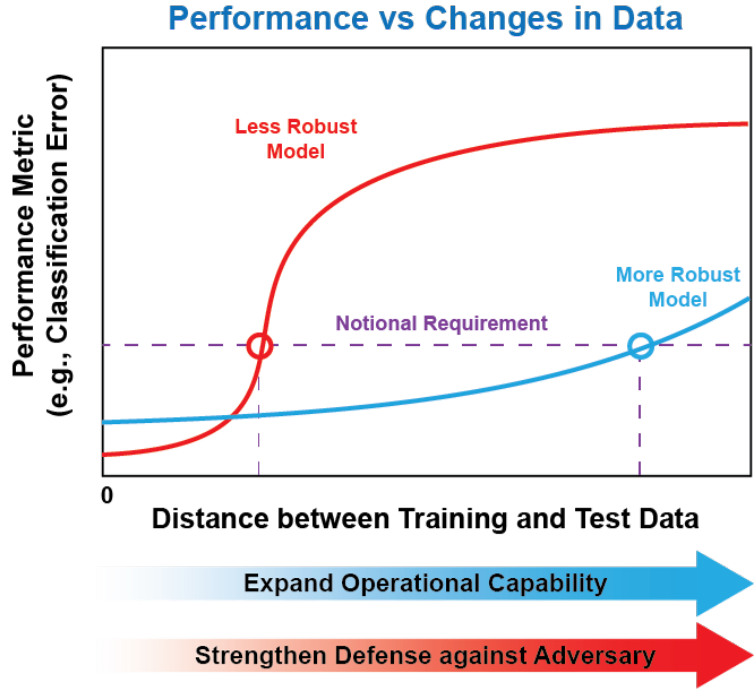

*Figure 5. Notional performance as a function of a distance measure between training and test data (e.g., size of perturbations applied to training samples or distance between original and out-of-distribution data) for a more robust (blue) and less robust (red) ML model. While the more robust model may see small increases in error rates on the original test set (distance = 0), it tolerates much larger changes in test data before error rates increase above the notional performance threshold. Robust models have an increased operational capability and present a greater challenge to potential adversaries.*

There are many different ways a set of data samples can be perturbed for robustness assessment. For example, a perturbation can be applied in a linear, additive manner (e.g., increasing or decreasing the values of individual image pixels), or by passing the input through a nonlinear transformation function (e.g., rotating or compressing the resolution of an image). A model evaluator may need to create multiple separate tests by perturbing different aspects of the data to get a good understanding of the model's overall sensitivity to perturbations. A few additional considerations when generating perturbed test sets include:

- **Perturbing Measurements vs. Real-World Phenomena**: Model inputs are typically measurements of some real-world phenomena collected by a sensor. Robustness of an AI/ML model can be assessed by perturbing those measurements, or it can be assessed at the system level by perturbing the real phenomena directly. For example in an audio system, a perturbation to the audio signal could be applied using Gaussian noise, while a perturbation to the real phenomena could be emulated by having a human speaker repeat the same phrase with varying levels of background noise. In many applications, it is easier and less costly to assess robustness to uncertainty in measurements compared to real phenomena. It may be difficult, however, to tie perturbations applied to measurements back to the underlying, real phenomena. Thus, when possible, performance should also be evaluated against perturbations to real phenomena.

- **Random vs. Worst-Case Perturbations**: Model-agnostic perturbations can be applied randomly to estimate a model's *average robustness* to common data corruptions such as sensor noise and environmental factors [5]. Alternatively, worst-case perturbations (i.e., those most likely to lead to performance degradation) for a specific model can be computed for each individual input via an optimization procedure [7]. Robustness curves computed using these optimized perturbations represent a lower bound or *worst-case robustness* for the model. This worst-case perturbation approach is becoming increasingly common in the AI/ML literature (often referred to as *adversarial robustness*), and is encouraged even when manipulation of inputs by an adversary is assumed to be unlikely, as it may reveal potential vulnerabilities of the model. Model evaluators should report performance given both average and worst-case perturbations.

- **Measure of Perturbation Size**: Creating robustness curves requires not only producing the perturbed test sets and computing performance on those sets, but also measuring the distance between the perturbed and unperturbed samples. The distance metric will vary depending on the type of data being perturbed and the perturbation method. For example, when assessing the sensitivity of a model to rotations of an input image, the distance measure is simply the angle of rotation that was applied to the image. For perturbations to image pixels, a common distance measure is the Euclidean norm. When perturbing real phenomena, specialized measures may need to be designed to represent the size of perturbations between targets of interest and varying environmental conditions.

Model evaluators will need to decide how to balance the cost of generating perturbed datasets with the valuable insight gained from assessing robustness to sample uncertainty. Upfront investment in such evaluations will enable evaluators to identify any vulnerabilities in the model before it is made operational and reduce the risk of dangerous and costly model failures in deployment. Additionally, assessing robustness to sample uncertainty is not only useful to model evaluators, but also to developers as they make design decisions. The DoD should encourage and, when appropriate, require robustness curves like those depicted in Figure 5 to be reported on during the evaluation phase, and should prioritize models that can tolerate larger perturbations before dropping below a desired performance threshold. Such models, with demonstrated robustness to sample uncertainty, will be less sensitive to noise and more difficult for an adversary to fool.

## 3.4 EVALUATION OF ROBUSTNESS TO DISTRIBUTION UNCERTAINTY

So far, we have discussed measures and recommendations for assessing performance of AI/ML models under the assumption that the data in the evaluation set is sampled from the same data distribution as the training set and small perturbations of those samples. When a model is deployed, the operational data is often from a slightly (or entirely) different distribution than the one assumed during model training. This *out-of-distribution data* may be due to a variety of factors, including incorrect assumptions made during the design and evaluation stage and unexpected changes in the scenario once the model is deployed.

To have a better understanding of how a model may perform given distribution uncertainty in the evaluation stage, the model evaluator should also attempt to estimate the model's performance given out-of-distribution data. A robust model should gracefully degrade in performance as the out-of-distribution data becomes increasingly different from the distribution assumed during the model design phase. Similar to the previous section, this robustness property can be assessed by creating robustness curves, where the horizonal axis of Figure 5 now reflects the distance between the original and out-of-distribution datasets. This distance can be computed using common distance metrics for distributions, such as the Kullback-Leibler divergence [8], or hand-crafted measures designed by domain experts.

To create a collection of test sets for such an assessment, the model evaluator needs to leverage domain expertise, intelligence information, and knowledge of the original training and testing data distribution to identify new datasets that represent realistic and feasible out-of-distribution data. For example, new distributions can be created using data from a new sensor, which may have different noise properties, or for a new threat that was not previously included in the training and testing datasets. Model evaluators also need to decide what level of performance to require on these more challenging datasets. For example, if the data in the new set is from sensors or threats that are significantly different from those assumed during model design, the accuracy requirement may be more lenient compared to a dataset that represents only a small deviation from the training set.

Assessing robustness given distribution uncertainty may require significant upfront investment to create the additional testing sets and define appropriate requirements and distance measures. While it may not be necessary for all applications, it is useful in any domain where the true data distribution remains uncertain or is likely to evolve over time. Such an assessment may reveal potential vulnerabilities in the design that can be fixed before the AI/ML model is deployed, and increase operator confidence that the system will perform as desired during the entirety of its operation. AI/ML models that are shown to perform well in the presence of distribution shift during the evaluation stage will likely require fewer updates during the deployment stage.

The principles and practices for DoD AI/ML model evaluators presented in the previous few sections are illustrated in Figure 6. The true and assumed data distributions (for training and generalization evaluation) are represented at the top. The bottom row depicts examples of how the different evaluation practices presented in this paper relate to the data distribution. Generalization is a measure of performance on new independently-drawn samples from the same data distribution (lower left). Robustness is a measure of performance degradation when perturbations are applied to individual data samples (lower middle) or when the test distribution shifts away from the assumed training distribution (lower right).

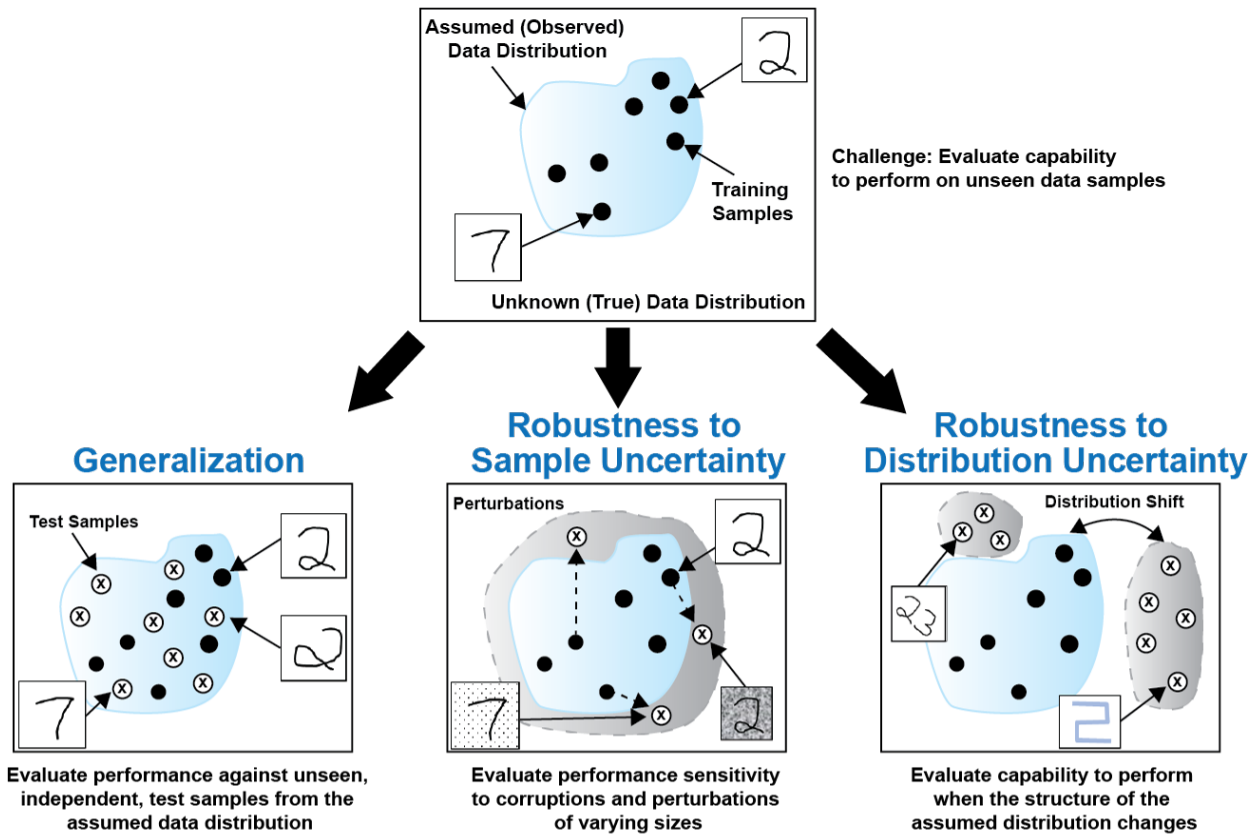

Figure 6. *Depiction of the best practices for AI/ML model evaluation presented in this paper. (Top) An illustration of the unknown, true data distribution and the assumed distribution from which training and test data are drawn, for an example AI/ML application of handwritten digit classification. The bottom row illustrates how the robustness properties outlined in this paper relate to the data distribution used for training and testing. (Lower left) Generalization is a measure of performance on unseen, or test data that is similar to the training data but independently sampled from the assumed distribution. Robustness is a measure of performance degradation when (lower middle) perturbations are applied to individual data samples, e.g., injecting noise to samples or applying adversarial perturbations, and when (lower right) the test distribution shifts away from the assumed distribution during training, e.g., black and white images to color.*

# 4. SUMMARY AND RECOMMENDATIONS

This paper provides an overview of the fundamental concepts needed for the DoD to improve existing processes to evaluate the performance and robustness of AI/ML models. There are three major phases of development for AI/ML: design of a model for a desired function, the verification of the model performance during evaluation, and transition to a deployed operational capability. Evaluation helps avoid deploying models that appear to meet a desired performance threshold but, in fact, have limited capability in operational settings. An effective evaluation process helps build trust between developers and operators in the field, discover important limitations of a given model for iterations of the design, and drive design of a more robust and resilient AI/ML model before deploying as an operational capability. Robust AI/ML models are more likely to maintain operational feasibility for longer periods of time without the need for costly updates.

Below we provide a list of recommended best practices for developing a process to evaluate models and drive designs toward more robust AI/ML capabilities. These recommendations are only a subset of possible approaches, but by implementing these recommended practices the DoD should be able to better estimate and improve the robustness of AI/ML models before deploying operationally. The recommendations are:

- **Develop large testing datasets with sufficient diversity to better represent the distribution of future, unseen data.** The true data distribution is not typically known, thus test datasets have inherent assumptions that can result in biased and potentially varying estimates of the model's true performance. Therefore, developing an appropriate sampling procedure to define the testing sets is essential to proper evaluation of the out-of-sample error.

- **Maintain separate testing and training datasets to ensure an honest and unbiased assessment against unseen data.** Keeping the test set independent of the design process helps limit the bias in the estimate of the model's true out-of-sample error. Given the iterative nature of the design and evaluation phase, vary the testing dataset (e.g., by randomizing) so the developer cannot utilize feedback on the test set performance to tune the model. Also, consider comparing the model developer's predicted performance based on the training/validation data during the design phase to the evaluated performance on the test data. A large gap between these values is an indication of possible overfitting of the training data. Requiring the developer to provide an estimate of the expected performance on the unseen test set encourages a more robust model design before the model is delivered to the model evaluator.

- **Assess a model's sensitivity to small input changes (corruptions and perturbations, random and worst-case).** A robust model should be able to maintain performance in the presence of small changes in the data samples and performance should gracefully degrade as those changes become larger. This type of assessment also verifies that the model has certain properties expected for a given application, such as invariance to rotation or scaling in image classification. It will be necessary to develop methods to generate these perturbations in order to perform these robustness tests.

- **Assess a model's capability against out-of-distribution data with respect to the training and test data.** The expectation is that the model maintains performance on data that is similar, but out-of-distribution, and degrades gracefully as the data appears more dissimilar. This out-of-distribution data may be due to a variety of factors, including having an insufficient representation of the true distribution due to incorrect assumptions or modeling challenges, or unexpected changes in the scenario. Whenever possible, evaluate the model on data from new threats, sensors, or environmental conditions to build confidence that the model will perform as expected and require fewer updates once deployed.

Model evaluators can motivate better design practices and create more robust systems by taking the proper approach to evaluation. Decisions about evaluation processes should be given careful consideration, and deviations from these best practices should be made only with a full understanding of the potential loss of capability and robustness. AI/ML evaluation processes in existing DoD programs should be assessed to determine any gaps between the status quo and the ideal. The next steps to address these gaps may depend on the application, in some cases requiring creative solutions depending on the status of current contracts. As evaluation processes improve, it is essential for the DoD to appropriately invest in tools and techniques that help model developers fully take advantage of the AI/ML revolution and devise robust solutions to our national security problems.